%% file: main.tex
\documentclass[runningheads]{llncs}
\usepackage{graphicx}
\usepackage{comment}
\usepackage{amsmath,amssymb} 
\usepackage{pifont}
\newcommand{\xmark}{\ding{55}}%
\usepackage{color}
\usepackage{bbm}
\usepackage{adjustbox}
\usepackage{tabularx}
\usepackage{breakcites}
\usepackage[width=122mm,left=12mm,paperwidth=146mm,height=193mm,top=12mm,paperheight=217mm]{geometry}

\DeclareRobustCommand\onedot{\futurelet\@let@token\@onedot}
\def\@onedot{\ifx\@let@token.\else.\null\fi\xspace}
\def\eg{\emph{e.g}.} 
\def\ie{\emph{i.e}.}

\def\etal{\emph{et al}.}

\makeatletter
\newcommand{\printfnsymbol}[1]{%
  \textsuperscript{\@fnsymbol{#1}}%
}
\makeatother

\begin{document}

\pagestyle{headings}
\mainmatter
\def\ECCVSubNumber{xxxx}  

\title{Self-supervised Keypoint Correspondences for Multi-Person Pose Estimation and Tracking in Videos}

\titlerunning{CorrTrack}
%
\author{Umer Rafi\inst{1}\thanks{equal contribution} \and
Andreas Doering\inst{1}\printfnsymbol{1} \and
Bastian Leibe\inst{2} \and Juergen Gall\inst{1}}
\authorrunning{U. Rafi et al.}
%
\institute{University of Bonn, Germany \and
RWTH Aachen, Germany}
\maketitle

\input{include/abstract}
\input{include/introduction}
\input{include/related_work}
%
\input{include/method_overview}
\input{include/correspondence_model}

\input{include/tracking}

\input{include/experiments}

\clearpage
\input{include/supplementary_material}

%
%
\bibliographystyle{splncs04}
\bibliography{egbib}

\end{document}

%% file: include/abstract.tex
\begin{abstract}
  Video annotation is expensive and time consuming. Consequently,
  datasets for multi-person pose estimation and tracking are less diverse and have more sparse annotations compared to large scale image datasets for human pose estimation.  
	This makes it challenging to learn deep learning based models for associating keypoints across frames that are robust to nuisance factors such as motion blur and occlusions for the task of multi-person pose tracking. 
To address this issue, we propose an approach that relies on keypoint correspondences for associating persons in videos. Instead of training the network for estimating keypoint correspondences on video data, it is trained on a large scale image dataset for human pose estimation using self-supervision. Combined with a top-down framework for human pose estimation, we use keypoint correspondences to (i) recover missed pose detections and to (ii) associate pose detections across video frames. Our approach achieves state-of-the-art results for multi-frame pose estimation and multi-person pose tracking on the PoseTrack 2017 and 2018 datasets. 
\end{abstract}

%% file: include/introduction.tex
\section{Introduction}

\begin{figure*}[t]
\begin{center}
 \includegraphics[width=\linewidth]{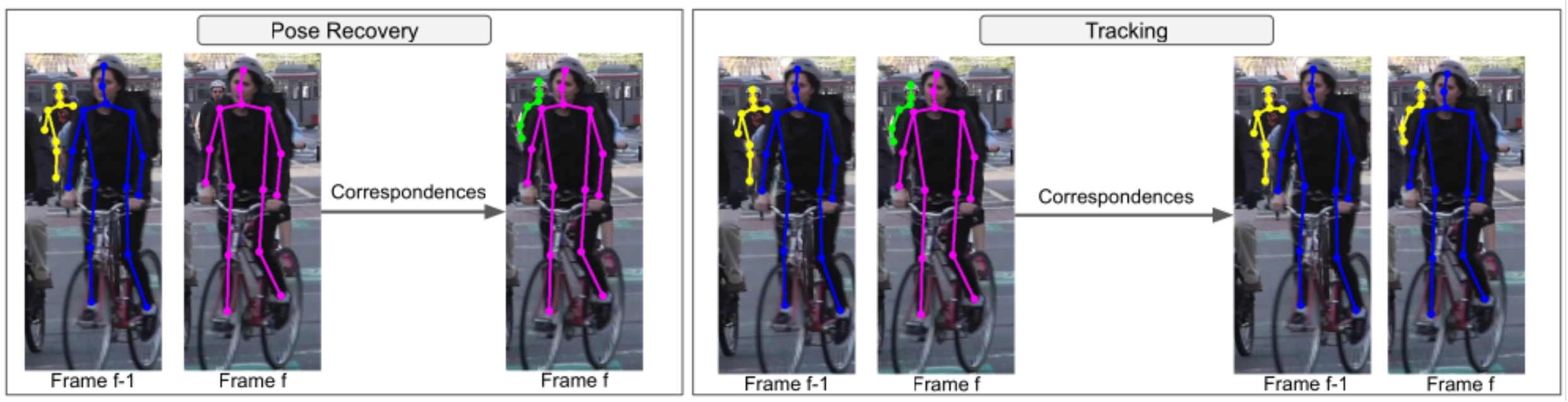}
\end{center}
 \caption{Our contributions: (Left) We use keypoint correspondences to recover missed pose detections by using the temporal context of the previous frame. (Right) We use keypoint correspondences to associate detected and recovered pose detections for the task of multi-person pose tracking.}
\label{fig:intro}
\end{figure*}

Human pose estimation is a very active research field in computer vision that is relevant for many applications like computer games, security, sports, and autonomous driving. Over the years, the human pose estimation models have been greatly improved~\cite{PSM,generative_partition_networks,cao2017realtime,kocabas18prn,simple_baseline,PoseWrapper,li2019rethinking} due to the availability of large scale image datasets for human pose estimation \cite{mscoco,mpii_dataset,aichallenger}. More recently, researchers started to tackle the more challenging problem of multi-person pose tracking \cite{Iqbal_2017_CVPR,insafutdinov17arttrack,simple_baseline,hrnet,offset_guided_networks}. 

In multi-person pose tracking, the goal is to estimate human poses in all frames of a video and associate them over time. However, video annotations are costly and time consuming. Consequently, recently proposed video datasets \cite{PoseTrack} are less diverse and are sparsely annotated as compared to large scale image datasets for human pose estimation \cite{mscoco,aichallenger}. This makes it challenging to learn deep networks for associating human keypoints across frames that are robust to nuisance factors such as motion blur, fast motions, and occlusions as they occur in videos.

State-of-the-art approaches \cite{simple_baseline,hrnet,MDPN} rely on optical flow or person re-identification \cite{offset_guided_networks} in order to track the persons. Both approaches, however, have disadvantages.   
Optical flow fails if a person becomes occluded which results in a lost track. While person re-identification allows to associate persons even if they disappeared for a long time, it remains difficult to associate partially occluded persons with person re-identification models that operate on bounding boxes of the full person. Moreover, the limited annotations in pose tracking datasets require to train the models on additional datasets for person re-identification. 

We therefore propose to learn a network that infers keypoint correspondences for multiple persons. The correspondence network comprises a Siamese matching module that takes a frame with estimated human poses as input and estimates the corresponding poses for a second frame. 
Such an approach has the advantage that it is not limited to a fixed temporal frame distance, and it allows to track persons when they are partially occluded. Our goal is to utilize keypoint correspondences to recover missed poses of a top-down human pose estimator, \eg, due to partial occlusion, and to utilize keypoint correspondences for multi-person tracking. 
The challenge, however, is to train such a network due to the sparsely annotated video datasets. In fact, in this work we consider the extreme case where the network is not trained on any video data or a dataset where identities of persons are annotated.          
Instead we show that such a network can be trained on an image dataset for multi-person pose estimation \cite{mscoco}. Besides of the human pose annotations, which are anyway needed to train the human pose estimator, the approach does not require any additional supervision. 
In order to improve the keypoint associations, we propose an additional refinement module that refines the affinity maps of the Siamese matching module.
 
\newpage
To summarize, the contributions of the paper are:
\begin{itemize}
\item We propose an approach for multi-frame pose estimation and multi-person pose tracking that relies on self-supervised keypoint correspondences which are learned from a large scale image dataset with human pose annotations. 
\item Combined with a top-down pose estimation framework, we use keypoint correspondences in two ways as illustrated in Figure \ref{fig:intro}: We use keypoint correspondences to (i) recover pose detections that have been missed by the top-down pose estimation framework and to (ii) associate detected and recovered poses in different frames of a video.
\item We evaluate the approach on the PoseTrack 2017 and 2018 datasets for the tasks of multi-frame pose estimation and multi-person pose tracking. Our approach achieves state-of-the-art results without using any additional training data except of \cite{mscoco} for the proposed correspondence network.   
 
\end{itemize} 

%% file: include/related_work.tex
\begin{table}[t]
\caption{Overview of related works on multi-person pose tracking.}\label{table:related_work}
	\centering
	\scalebox{0.90}{\begin{tabular}{ccc}
		\hline 
		Method  & Detection Improvement & Tracking \\
		\hline
		Ours & Correspondences & Keypoint Correspondences \\ 
		HRNet \cite{hrnet}  & Temporal OKS & Optical Flow \\
		POINet \cite{poinet} & - & Ovonic Insight Net \\
		MDPN \cite{MDPN} & Ensemble & Optical FLow \\ 
		LightTrack \cite{ning2019lighttrack} & Ensemble / BBox Prop. & GCN \\
		ProTracker \cite{detect_n_track} & - & IoU \\
		\hline
		STAF \cite{staf} &  - & ST Fields \\
		ST Embeddings \cite{stembedding}& - & ST Embeddings \\
		JointFlow \cite{jointflow}&  - & Flow Fields
	\end{tabular}}
	
\end{table}

\section{Related Work}
\paragraph{Multi-Person Pose Estimation.}
Multi-person pose estimation can be categorized into top-down and bottom-up approaches. Bottom-up based methods \cite{kocabas18prn,cao2017realtime,associative_embedding,mask_rcnn,generative_partition_networks} first detect all person keypoints simultaneously and then associate them to their corresponding person instances.
For example, Chao \etal~\cite{cao2017realtime} predict part affinity fields which provide information about the location and orientation of the limbs. For the association, a greedy approach is used. More recently, Kocabas \etal~\cite{kocabas18prn} propose to detect bounding boxes and pose keypoints within the same neural network. In the first stage, bounding box predictions are used to crop from predicted keypoint heatmaps. As a second stage, a pose residual module is proposed, which regresses the respective keypoint locations of each person instance.

Top-down methods \cite{learning_feature_pyramid,li2019rethinking,cascaded_pyramid_network,multi_scale_aggregation,simple_baseline,learning_feature_pyramid,stackedhourglass} utilize person detectors and estimate the pose on each image crop individually. In contrast to bottom-up methods, top-down approaches do not suffer from scale variations.

For example, Xiao \etal~\cite{simple_baseline} propose a simple yet strong model based on a ResNet152 \cite{he2015residual} and achieve state-of-the-art performance by replacing the last fully connected layer by three transposed convolutions.
Li \etal~\cite{li2019rethinking} propose an information propagation procedure within a multi-stage architecture based on four ResNet50 networks \cite{he2015residual} with coarse-to-fine supervision.

\paragraph{Multi-Frame Pose Estimation.}
In video data, such as PoseTrack \cite{PoseTrack}, related works \cite{PoseFlow,MDPN,PoseWrapper} leverage temporal information of neighboring frames to increase robustness against fast motions, occlusion, and motion blur. Xiu \etal~\cite{PoseFlow} and Guo \etal~\cite{MDPN} utilize optical flow to warp preceding frames into the current frame.
Recently, Bertasius \etal~\cite{PoseWrapper} propose a feature warping method based on deformable convolutions to warp pose heatmaps from preceding and subsequent frames into the current frame. While they show that they are able to learn from sparse video annotations, they do not address multi-person pose tracking. 

\paragraph{Multi-Person Pose Tracking.} 
Early works for multi-person pose tracking \cite{Iqbal_2017_CVPR,insafutdinov17arttrack} build spatio-temporal graphs which are solved by integer linear programming. Since such approaches are computationally expensive, researchers reduced the task to bipartite graphs which are solved in a greedy fashion \cite{hrnet,poinet,MDPN,ning2019lighttrack,detect_n_track,jointflow,staf,PoseFlow,stembedding}. Girdhar \etal~\cite{detect_n_track} propose a 3D Mask R-CNN \cite{mask_rcnn} to generate person tubelets for tracking which are associated greedily.
 
More recent works \cite{simple_baseline,hrnet,MDPN,offset_guided_networks} incorporate temporal information by using optical flow. Xiao \etal~\cite{simple_baseline} rely on optical flow to recover missed person detections and propose an optical-flow based similarity metric for tracking. In contrast, Zhang \etal~\cite{offset_guided_networks} builds on \cite{detect_n_track} and propose an adapted Mask R-CNN \cite{mask_rcnn} with a greedy bounding box generation strategy. Furthermore, optical flow and a person re-identification module are combined for tracking. 
Jin \etal~\cite{stembedding} perform multi-person pose estimation and tracking within a unified framework based on human pose embeddings. Table \ref{table:related_work} provides a summary of the contributions of the recent related works.

\paragraph{Correspondences.} In recent years, deep learning has been successfully applied to the task of correspondence matching \cite{UCN,FCSS,LSC}, including the task of visual object tracking (VOT) \cite{bertinetto2016fully,Li_2018_CVPR,vot2,vot3}. All of the above approaches establish correspondences at object level. In contrast, our approach establishes correspondences at instance level. Moreover, VOT tasks assume a ground truth object location for the first frame, which is in contrast to the task of pose tracking.

\paragraph{Self-Supervised Learning.} Self-supervised learning approaches \cite{CycleTime,uvc_2019} have been proposed for establishing correspondences at patch and keypoints level from videos. However, these approaches use videos for learning and process a single set of keypoints or patch at a time. In contrast, our approach establishes correspondences for multiple instances and is trained on single images. 

%% file: include/method_overview.tex
\begin{figure*}[t]
\begin{center}
 \includegraphics[width=0.98\linewidth]{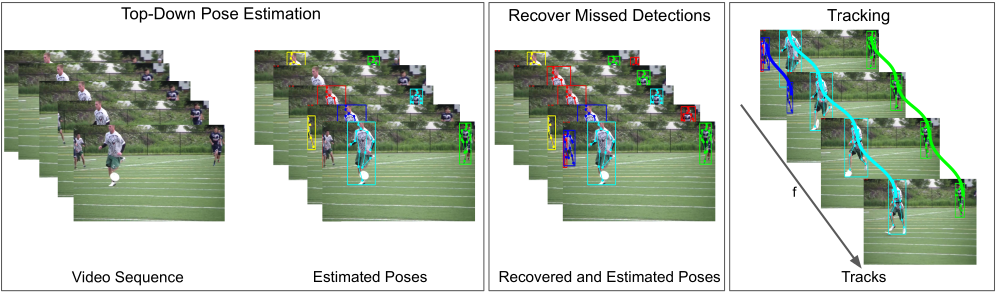}
\end{center}
   \caption{Given a sequence of frames, we detect a set of person bounding boxes and perform top-down pose estimation. Our proposed method uses keypoint correspondences to (i) recover missed detections and (ii) to associate detected and recovered poses to perform tracking. The entire framework does not require any video data for training since the network for estimating keypoint correspondences is trained on single images using self-supervision.}
\label{fig:overview}
\end{figure*}

\section{Method Overview}\label{sec:method}
In this work, we propose a multi-person pose tracking framework that is robust to motion blur and severe occlusions, even though it does not need any video data for training.    
As it is illustrated in Figure \ref{fig:overview}, we first estimate for each frame the human poses and then track them.   
 
For multi-person human pose estimation, we utilize an off-the-shelf object detector \cite{cascaded_RCNN} to obtain a set of bounding boxes for the persons in each frame. For each bounding box, we then perform multi-person pose estimation in a top-down fashion by training an adapted GoogleNet \cite{googlenet}, which we will discuss in Section \ref{sec:implementation_details}. 

In order to be robust to motion blur and severe occlusions, we do not use optical flow in contrast to previous works like~\cite{simple_baseline}. Instead we propose a network that estimates for a given frame with estimated keypoints the locations of the keypoints in another frame. We use this network for recovering human poses that have been missed by the top-down pose estimation framework as described in Section~\ref{sec:pose_propagation} and for associating detected and recovered poses across the video as described in Section~\ref{sec:frame_by_frame}. 

The main challenge for the keypoint correspondence network is the handling of occluded keypoints and the limited amount of densely annotated video data. In order to address these issues, we do not train the network on video data, but on single images using self-supervision. In this way, we can simulate disappearing keypoints by truncation and leverage large scale image dataset like MS-COCO \cite{mscoco} for tracking. We will first describe the keypoint correspondence network in Section \ref{sec:corr_framework} and then discuss the tracking framework in Section \ref{sec:tracking_approach}.   

%% file: include/correspondence_model.tex
\begin{figure*}
\begin{center}
 \includegraphics[width=\linewidth]{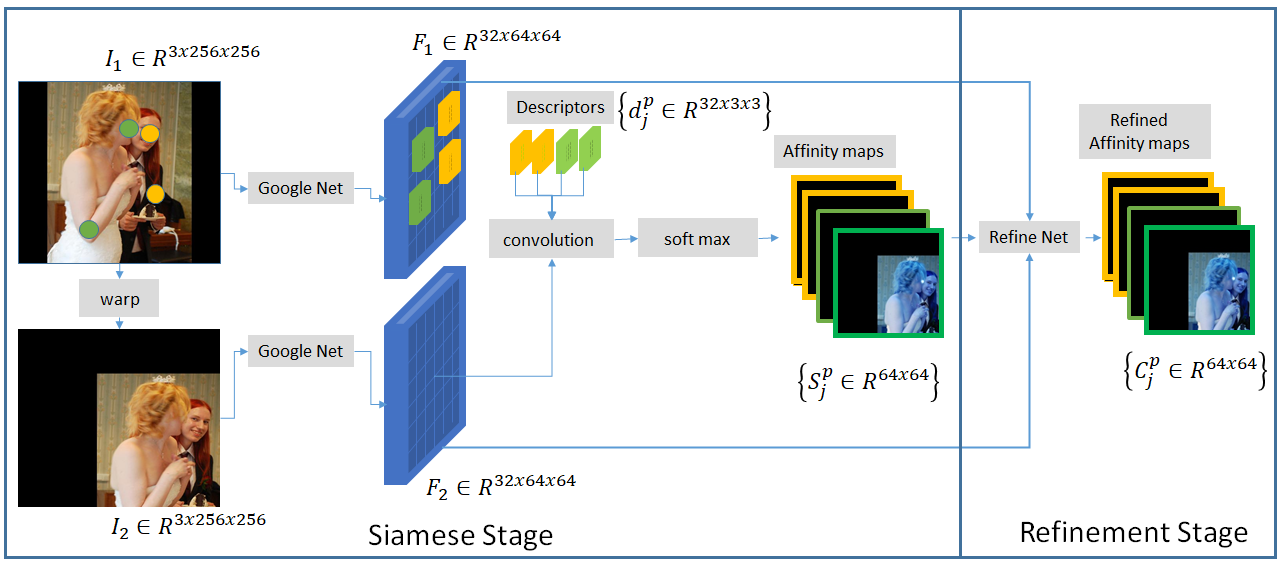}
\end{center}
   \caption{Keypoint correspondence network. The Siamese network takes images $I_1$ and $I_2$ and keypoints $\{j^p\}_{1:N_p}$ for all persons $p$ in image $I_1$ as input and generates the feature maps $F_1$ and $F_2$, respectively. The keypoints of the different persons are shown in green and yellow, respectively. 
	For each keypoint, a descriptor $d^p_j$ is extracted from $F_1$ and convolved with the feature map $F_2$ to generate an affinity map $S^p_j$. In order to improve the affinity maps for each person, the refinement network takes $F_1$, $F_2$ and the affinity maps $S^p_j$ for person $p$ as input and generates refined affinity maps $C^p_j$.}
\label{fig:CN}
\end{figure*}
\section{Keypoint Correspondence Network}\label{sec:corr_framework}

Given two images $I_1$ and $I_2$ with keypoints $\{j^p\}_{1:N_p}$ for all persons $p$ in image $I_1$, our goal is to find the corresponding keypoints in $I_2$. Towards this end, we use a Siamese network as shown in Figure \ref{fig:CN} which estimates for each keypoint an affinity map. The affinity maps are further improved by the refinement module, which is described in Section~\ref{sec:refinement_module}.
 
\subsection{Siamese Matching Module}
\label{DS}

The keypoint correspondence network consists of a Siamese network. Each branch in the Siamese network is a batch normalized GoogleNet up to layer 17 with shared parameters \cite{googlenet}. The Siamese network takes an image pair $(I_1$, $I_2)$ and keypoints $\{j^p\}_{1:N_p}$ for persons $p \in \{1,\dots,P\}$ in the image $I_1$ as input. During training, $I_2$ is generated by applying a randomly sampled affine warp to $I_1$. In this way, we do not need any annotated correspondences during training or pairs of images, but train the network on single images with annotated poses. We use an image resolution of $256\times256$ for both images.

The Siamese network generates features $F_1 \in \mathbb{R}^{32\times64\times64}$ and $F_2 \in \mathbb{R}^{32\times64\times64}$ for images $I_1$ and $I_2$, respectively. The features are then pixel-wise $l_2$ normalized and local descriptors $d^p_j \in R^{32\times3\times3}$ are generated for each keypoint $j^p$ by extracting squared patches around the spatial position of a keypoint in the feature maps $F_1$.

Given a local descriptor $d^p_j$, we compute its affinity map $A^p_j$ over all pixels $x=\{1,\dots,64\}$ and $y=\{1,\dots,64\}$ in $F_2$  as:
\begin{equation}
A^p_j = d^p_j \circledast F_2
\end{equation}
where $\circledast$ denotes the convolution operation. 
Finally, a softmax operation is applied to the affinity map $A^p_j$, \ie,
\begin{equation}
S^p_j(x, y)  = \frac{\exp(A^p_j(x, y))}{\sum_{x',y'}\exp(A^p_j(x', y'))}.
\end{equation}
We refine the affinity maps $S^p_j$ further using a refinement module.

\subsection{Refinement Module}\label{sec:refinement_module}
Similar to related multi-stage approaches \cite{cao2017realtime,Reference1,kocabas18prn}, we append a second module to the keypoint correspondence network to improve the affinity maps generated by the Siamese matching module. For the refinement module, we use a batch normalized GoogleNet from layer 3 till layer 17. The refinement module concatenates $F_1$,  $F_2$, and the affinity maps $\{S^p_j\}_{1:N_p}$ for a single person $p$ and refines the affinity maps, which we denote by $C^p_j \in \mathbb{R}^{64\times64}$. The refinement module is therefore applied to the affinity maps for all persons $p \in \{1,\dots,P\}$. Before we describe in Section~\ref{sec:tracking_approach} how we will use the affinity maps for tracking $C^p_j$, we describe how the keypoint correspondence network is trained.

\subsection{Training}
\label{LO}

Since we train our network using self-supervision, we train it using a single image $I_1$ with annotated poses. We generate a second image $I_2$ by applying a randomly sampled affine warp to $I_1$. We then generate the ground-truth affinity map $G^p_j$ for a keypoint $j^p$ belonging to person $p$ as:
\begin{equation}
  G^p_j(x, y) = 
\begin{cases}
    1 & \text{if $x=\hat{x}^p_j$ and $y=\hat{y}^p_j$},\\
    0 & \text{otherwise},
  \end{cases}
\end{equation}
where $(\hat{x}^p_j, \hat{y}^p_j)$ is the spatial position of the ground-truth correspondence for keypoint $j^p$ in image $I_2$, which we know from the affine transformation. As illustrated in Figure~\ref{fig:CN}, not all corresponding keypoints are present in image $I_2$. In this case, the ground-truth affinity map is zero and predicting a corresponding keypoint is therefore penalized.

During training, we minimize the binary cross entropy loss between the predicted affinity maps $S^p_j$ and $C^p_j$ and the ground-truth affinity map $G^p_j$: 
\begin{align}
&\min\limits_{\theta}\sum_{x,y}-\left(G^p_j\log(S^p_j) + (1-G^p_j)(1-\log(S^p_j)\right),\\
&\min\limits_{\theta}\sum_{x,y}-\left(G^p_j\log(C^p_j) + (1-G^p_j)(1-\log(C^p_j)\right),
\end{align}
where $\theta$ are the parameters of the keypoint correspondence framework.

%% file: include/tracking.tex
\section{Multi-Person Pose Tracking}\label{sec:tracking_approach}

\begin{figure}[t]
\begin{center}
 \includegraphics[width=0.9\linewidth]{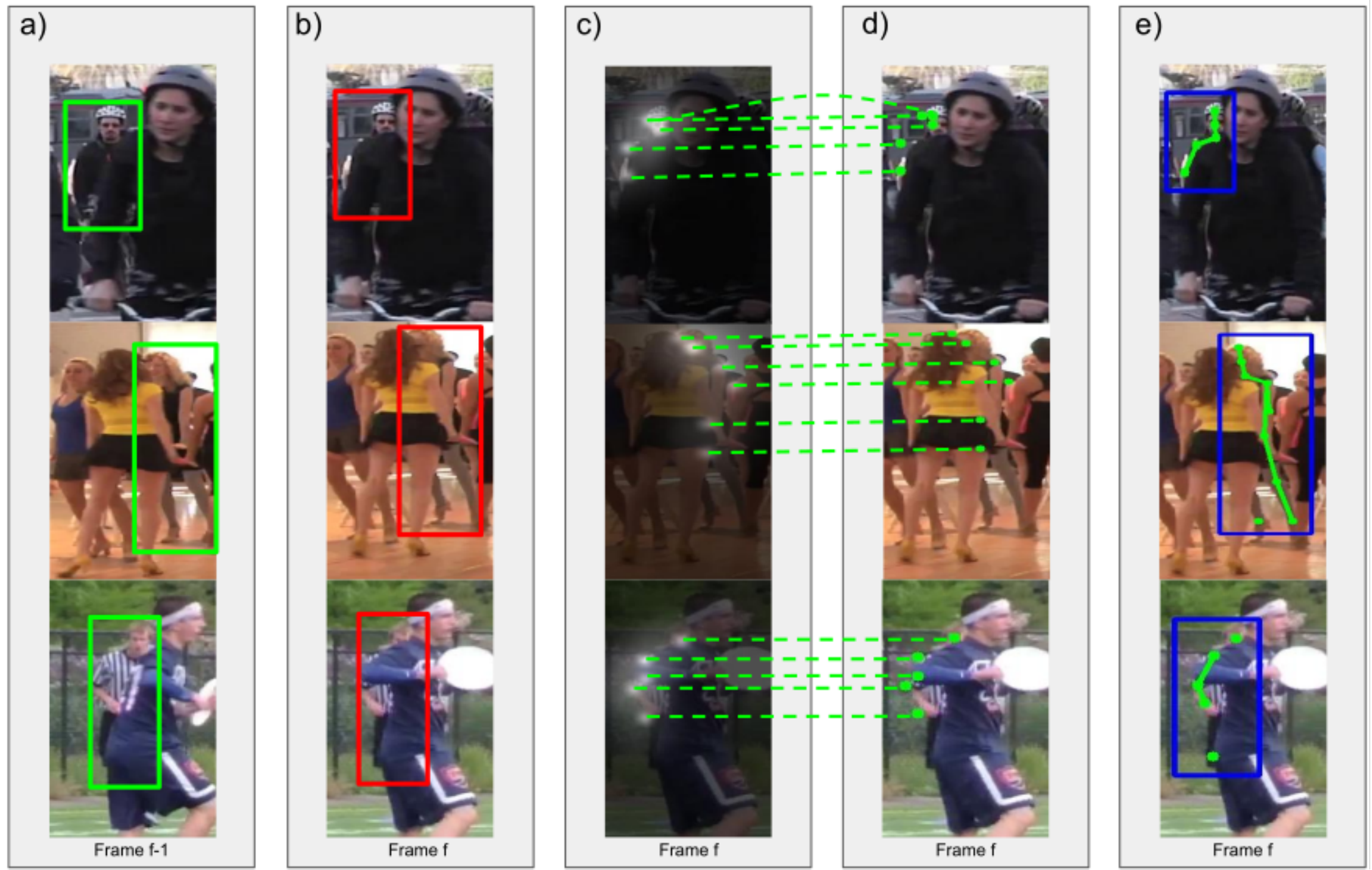}
\end{center}
 \caption{ Recovering missed detections. (a) Person detected by the top-down pose estimation framework in frame $f{-}1$. (b) Person missed by the top down pose estimation framework in frame $f$ due to occlusion. (c) Keypoint affinity maps of the missed person from frame $f{-}1$  to frame $f$. (d) Corresponding keypoints in frame $f$. (e) Estimated bounding box from the corresponding keypoints and the recovered pose. }
\label{fig:pose_recovery}
\end{figure}    

\begin{figure}[h]
\begin{center}
 \includegraphics[width=\linewidth]{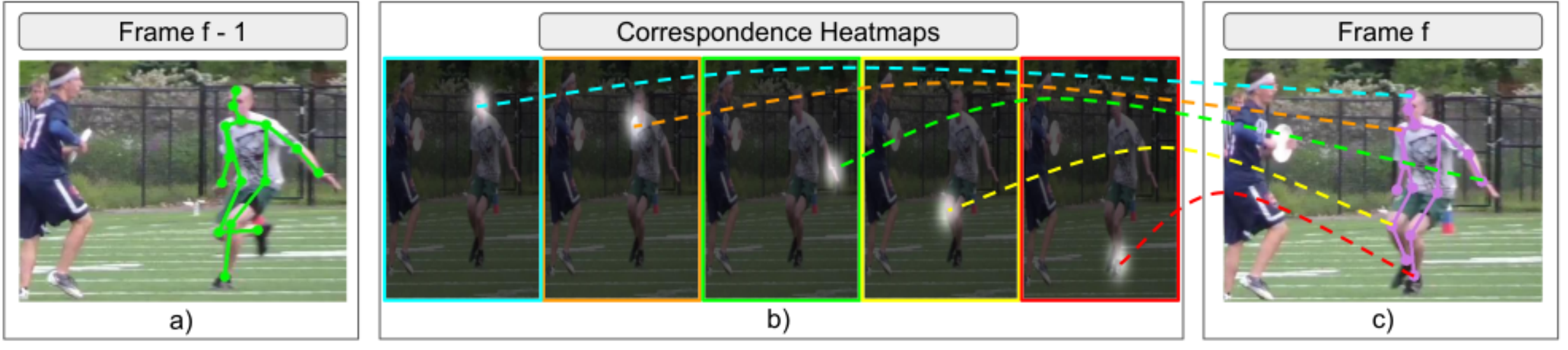}
\end{center}
 \caption{ Pose to track association. (a) A tracked human pose till frame $f{-}1$. (b) Keypoint affinity maps of the track to frame $f$. (c) A pose instance of frame f. The dashed lines indicate the position of each detected joint of the pose instance in the correspondence affinity maps of the tracked pose in frame $f-1$.}
\label{fig:sim_corr}
\end{figure}

We use the keypoint correspondence network in two ways. First, we use it to recover human poses that have been missed by the frame-wise top-down multi-person pose estimation step, which will be described in Section \ref{sec:pose_propagation}. Second, we use keypoint correspondences for tracking poses across frames of the video as described in Section \ref{sec:frame_by_frame}.

\subsection{Recover Missed Detections}\label{sec:pose_propagation}

For a given frame $f$, we first detect the human poses in the frame using the top-down multi-person pose estimator described in Section~\ref{sec:implementation_details}. While the person detector \cite{cascaded_RCNN} performs well, it fails in situations with overlapping persons and motion blur. Consequently, the human pose is not estimated in these cases.  Examples are shown in Figure \ref{fig:pose_recovery}(b). 

Given the detected human poses $J^p_{f-1} = \{{j}^p_{f-1}\}$ for persons $p \in \{1,\dots,P\}$ in frame $f{-}1$, we compute the corresponding refined affinity maps $C^p = \lbrace C_j^p\rbrace$ by using the keypoint correspondence network. For each keypoint $j^p_{f-1}$, we then get the corresponding keypoint $\bar{j}^p_{f}$ in frame $f$ by taking the argmax of $C_j^p$ and mapping it to the image resolution. Since the resolution of the affinity maps is lower than the image resolution and since the frame $f$ might contain a keypoint that was occluded in the previous frame, we reestimate the propagated poses.
This is done by computing for each person $p$ a bounding box that encloses all keypoints $\bar{J}^p_{f} = \{\bar{j}^p_{f}\}$ and using the human pose estimation network described in Section~\ref{sec:implementation_details} to get a new pose for this bounding box. We denote the newly estimated poses by $\hat{J}^p_{f}$. The overall procedure is shown in Figure \ref{fig:pose_recovery}. We apply OKS based non-maximum suppression \cite{simple_baseline} to discard redundant poses.

\subsection{Tracking}\label{sec:frame_by_frame}
Given detected and recovered poses, we need to link them across video frames to obtain tracks of human poses. Tracking can be seen as a data association problem  over estimated poses. Previously, the problem has been approached using bipartite graph matching \cite{detect_n_track} or greedy approaches \cite{simple_baseline,hrnet,jointflow}. In this work, we greedily associate estimated poses over time by using the keypoint correspondences. We initialize tracks on the first frame and then associate new candidate poses to intial tracks one frame at a time. 

Formally, our goal is to assign pose instances $\{B^{p}_{f}\}=\{J^{p}_{f}\} \cup \{\hat{J}^{p}_{f}\}$ in frame $f$ for persons $p \in \lbrace 1, \ldots P \rbrace$ to tracks $\{T^{q}_{f{-}1}\}$ till frame $f{-}1$ for persons $ q \in \lbrace 1, \ldots Q \rbrace$. Towards this end, we measure the similarity between a pose instance $B^{p}_{f}$
and a track $T^{q}_{f{-}1}$ as:

\begin{equation}\label{eq:lookup}
		   S(T^{q}_{f-1}, B^p_f) = \frac{\sum_{j=1}^{N_q}C^{q}_j(j^p_f) \cdot \mathbbm{I}_{C^{q}_j(j^p_f) > \tau_{corr}}}{\sum_{j=1}^{N_q} \mathbbm{I}_{C^{q}_j(j^p_f) > \tau_{corr}}},
\end{equation}
where $C^{q}_j$ is the affinity map of the keypoint $j$ in track $T_{f-1}^{q}$ for frame $f$. The affinity map is computed by the network described in Section \ref{sec:corr_framework}.
$C^{q}_j(j^p_f)$ is the confidence value in the affinity map $C^{q}_j$ at the location of the joint $j^p_f$ for person $p$ in frame $f$. $N_q$ is the number of detected joints. An example is shown in Figure \ref{fig:sim_corr}. We only consider $j^p_f$ if its affinity is above $\tau_{corr}$. If a pose $B^{p}_{f}$ cannot be matched to a track $T^{q}_{f{-}1}$, a new track is initiated.

%% file: include/experiments.tex
\section{Experiments and Results}
We evaluate our approach on the Posetrack 2017 and 2018 datasets \cite{PoseTrack}. The datasets have 292 and 593 videos for training and 214 and 375 videos for evaluation, respectively. We evaluate multi-frame pose estimation and tracking results using the mAP and MOTA evaluation metrics.

\subsection{Implementation Details}
\label{sec:implementation_details}
We provide additional implementation details for our top-down pose estimation and keypoint correspondence network below.

\paragraph{Top-down Pose Estimation.}
We use a top-down framework for frame level pose estimation.
We use cascade R-CNN \cite{cascaded_RCNN} for person detection and extract crops of size $384 \times 288$ around detected persons as input to our pose estimation framework, which consists of two stages.
Each stage is a batch normalized GoogleNet \cite{googlenet}. The backbone in the first stage consists of layer 1 to layer 17 while the second backbone consists of layer 3 to layer 17 only. Both stages predict pose heatmaps and joint offset maps for the cropped person as in \cite{offset_guided_networks}. We use the pose heatmaps in combination with the joint offsets from the second stage as our pose detections.
The number of parameters (39.5 M) of our model is significantly lower compared to related works such as FlowTrack \cite{simple_baseline} (63.6 M) or EOGN \cite{offset_guided_networks} (60.3 M).

We train the pose estimation framework on the MS-COCO dataset \cite{mscoco} for $260$ epochs with a base learning rate of $1e^{-3}$. The learning rate is reduced to $1e^{-4}$ after $200$ epochs. During training we apply random flippings and rotations to input crops.
We finetune the pose estimation framework on the PoseTrack 2017 dataset \cite{PoseTrack} for 12 epochs. The learning rate is further reduced  to $1e^{-5}$ after epoch 7.

\paragraph{Keypoint Correspondence Network.}
We perform module-wise training. We first train the Siamese module. We then fix the Siamese module and train the refinement module. Both modules are trained for $100$ epochs with base learning rate of $1e^{-4}$ reduced to $1e^{-5}$ after $50$ epochs. We generate a second image for each training image by applying random translations, rotations, and flippings to the first image. The keypoint correspondence network is trained only on the MS-COCO dataset \cite{mscoco}. We did not observe any improvements in our tracking results by finetuning the correspondence model on the PoseTrack dataset. Training only on the PoseTrack dataset yielded sub-optimal tracking results since PoseTrack is sparsely annotated and contains far less person instances than MS-COCO.

\setlength{\tabcolsep}{4pt}
\begin{table}[t]
\caption{Comparison with tracking baselines on the PoseTrack 2017 validation set. For the comparison, the detected poses based on ground-truth bounding boxes (GT Boxes) or detected bounding boxes are the same for each approach. Correspondence based tracking consistently improves MOTA compared to the baselines and significantly reduces the number of identity switches (IDSW).}
	\label{table:baselines}
	\centering
	\scalebox{0.9}{\begin{tabular}{lcll}
		\hline\noalign{\smallskip}
		 Tracking Method & GT Boxes  & IDSW & MOTA \\
		\noalign{\smallskip}
		\hline
		\noalign{\smallskip}
		\hline
		OKS & \checkmark & 6582 & 65.9 \\
		Optical Flow  & \checkmark & 4419 & 68.4 \\
		Re-ID & \checkmark & 4164 & 67.1 \\
		Correspondences & \checkmark & 3583 & 70.5 \\
		\hline  
		OKS & \xmark  & 7207 & 60.4  \\ 
		Optical Flow & \xmark  &  5611 & 66.7 \\
		Re-ID  & \xmark & 4589 & 64.1 \\
		Correspondences  & \xmark & 3632 & 67.9 \\
	\end{tabular}}
	
\end{table}

\subsection{Baselines}\label{sec:ablation_studies}
We compare our keypoint correspondence tracking to different standard tracking baselines for multi-person pose tracking  as reported in Table \ref{table:baselines}. To measure the performance of each baseline, we report the number of identity switches and the MOTA score. For a fair comparison, we replace the keypoint correspondences in our framework by different baselines. For all experiments, we use the same detected poses using either ground truth or detected bounding boxes.

\textit{OKS.} OKS without taking the motion of the poses into account has been proposed in \cite{simple_baseline}. OKS measures the similarity between two poses and is independent of their appearance. It is not robust to large motion, occlusion, and large temporal offsets. This is reflected in Table \ref{table:baselines} as this baseline achieves the lowest performance.

\textit{Optical Flow.}
Optical flow is a temporal baseline that has been proposed in \cite{simple_baseline}. We use optical flow to warp the poses from the previous frame to the current frame. We then apply OKS for associating the warped poses with candidate poses in the current frame. We use the pre-trained PWC-net \cite{pwcnet} as
done in [37] for a fair comparison. Optical flow clearly outperforms OKS and achieves superior MOTA of 68.4 and 66.7 for GT and detected bounding boxes, respectively.

\textit{Person Re-id.}
Compared to optical flow and OKS, person re-identification is more robust to larger temporal offsets and large motion. However, the achieved results indicate that person re-identification operating on the bounding boxes performs sub-optimally under the frequent partial occlusions in the PoseTrack datasets. For our experiments, we use the pre-trained re-identification model from \cite{Luo_2019_CVPR_Workshops}. Re-identification based tracking achieves MOTA scores of 67.1 and 64.1 for GT and detected bounding boxes, respectively. 

The results show that correspondence based tracking (1) achieves a consistent improvement over the baselines for ground-truth and detected bounding boxes with MOTA scores of 70.5 and 67.9, respectively, and (2) significantly reduces the number of identity switches. Compared to optical flow, correspondences are more robust to partial occlusions, motion blur, and large motions. A qualitative comparison is provided in Section \ref{sec:qualitative_results}.

\subsection{Effect of Joint Detection Threshold and Pose Recovery}
\setlength{\tabcolsep}{4pt}
\begin{table}[t]
   \caption{Effect of joint detection threshold and pose recovery on mAP and MOTA for the PoseTrack 2018 validation set. The results are shown for (left) detected poses only and (right) detected and recovered poses. As expected, recovering missed detections improves both MOTA and mAP. A good trade-off between mAP and MOTA is achieved by the joint detection threshold $0.3$.
   }\label{table:joint_threshold_tracking}
    \begin{minipage}[t]{.5\linewidth}
      \centering
       \scalebox{0.85}{\begin{tabular}{lll}
		\hline
		 Joint Threshold & mAP & MOTA\\
		\hline
		\multicolumn{3}{c}{Detected Poses Only}\\
		\hline
		0.0 & 80.1 & 48.1 \\
		0.1 & 79.7 & 63.3 \\
		0.2 & 78.9 & 66.1 \\
		\textbf{0.3} & \textbf{77.7} & \textbf{67.6} \\
		0.4 & 75.9 & 68.0 \\
		0.5 & 73.1 & 67.1 \\
	\end{tabular}}
    \end{minipage}\hfill%
    \begin{minipage}[t]{.5\linewidth}
      \centering
       \scalebox{0.85}{\begin{tabular}{lll}
		\hline
		 Joint Threshold & mAP & MOTA\\
		\hline
		\multicolumn{3}{c}{Detected and Recovered Poses.} \\
		\hline
		0 & 82.0 & 48.1\\ 
		0.1 & 81.4 & 64.1 \\
		0.2 & 80.5 & 67.2 \\
		\textbf{0.3} & \textbf{79.2} & \textbf{68.8} \\
		0.4 & 77.2 & 69.2 \\
		0.5 & 74.2 & 68.2 \\
	\end{tabular}}
    \end{minipage} 
\end{table}
\noindent
We evaluate the impact of different joint detection thresholds on mAP and MOTA for the PoseTrack 2018 dataset as shown in Table \ref{table:joint_threshold_tracking}. Since mAP does not penalize false-positive keypoints, thresholding decreases the pose estimation performance by discarding low confident joints. Vice versa, joint thresholding results in cleaner tracks and improves the tracking performance, as MOTA penalizes false-positive keypoint detections. 
A good trade-off  between mAP and MOTA is achieved for the joint detection threshold $0.3$ resulting in mAP and MOTA of $77.7$ and $67.6$, respectively. 

While on the left hand side of Table \ref{table:joint_threshold_tracking} we report the results without recovering missed detections as described in Section \ref{sec:pose_propagation}, the table on the right hand side shows the impact on mAP and MOTA if missed detections are recovered. The recovering of missed detections improves the accuracy for all thresholds. For the joint detection threshold $0.3$, mAP and MOTA are further improved to $79.2$ and $68.8$, respectively.

\setlength{\tabcolsep}{4pt}
\begin{table}[t]
\caption{Comparison to the state-of-the-art on the PoseTrack 2017 and 2018 validation set for multi-frame pose estimation.
}\label{table:multi_frame_pose_sota}
	\centering
	 \scalebox{0.75}{\begin{tabular}{lllllllll|l}
		\hline\noalign{\smallskip}
		 Dataset & Method & Head & Shoulder & Elbow & Wrist & Hip & Knee & Ankle & mAP\\
		\noalign{\smallskip}
		\hline
		\noalign{\smallskip}
PoseTrack 17 Val Set & DetectNTrack \cite{detect_n_track} & 72.8 & 75.6 & 65.3 & 54.3 & 63.5 & 60.9 & 51.8 & 64.1 \\
& PoseFlow \cite{PoseFlow} & 66.7 & 73.3 & 68.3 & 61.1 & 67.5 & 67.0 & 61.3 & 66.5 \\
& FlowTrack \cite{simple_baseline} & 81.7 & 83.4 & 80.0 & 72.4 & 75.3 & 74.8 & 67.1 & 76.7 \\
& HRNet \cite{hrnet} & 82.1 & 83.6 & 80.4 & 73.3 & 75.5 & 75.3 & 68.5 & 77.3 \\
& MDPN \cite{MDPN} & 85.2 & 88.5 & 83.9 & 78.0 & 82.4 & 80.5 & 73.6 & 80.7 \\
& PoseWarper \cite{PoseWrapper} & 81.4 & 88.3 & 83.9 & 78.0 & 82.4 & 80.5 & 73.6 & \textbf{81.2} \\
& Ours & 86.1 & 87.0 & 83.4 & 76.4 & 77.3 & 79.2 & 73.3 & 80.8 \\
\hline
PoseTrack 18 Val Set & PoseFlow \cite{PoseFlow} &63.9 & 78.7 & 77.4 & 71.0 & 73.7 & 73.0 & 69.7 & 71.9 \\
		&MDPN \cite{MDPN} & 75.4 & 81.2 & 79.0 & 74.1 & 72.4 & 73.0 & 69.9 & 75.0 \\
		& PoseWarper \cite{PoseWrapper} & 79.9 & 86.3 & 82.4 & 77.5 & 79.8 & 78.8 & 73.2 & 79.7 \\
		&Ours  &86.0 &87.3 &84.8 & 78.3 & 79.1 & 81.1 & 75.6 & \textbf{82.0} 
	\end{tabular}}
	
\end{table}
\setlength{\tabcolsep}{1.4pt}

\subsection{Comparison with State-of-the-art Methods}
\begin{table}[t]
\caption{Comparison to the state-of-the-art on the PoseTrack 2017 and 2018 validation and test sets. Approaches marked with $^+$ use additional external training data. Approaches marked with $^*$ do not report results on the official test set}\label{table:state-of-the-art-tracking}
\begin{minipage}[t]{.49\linewidth}
      \centering
        \scalebox{0.70}{\begin{tabular}{l|llll}
        \hline
		 &  Approach & mAP & MOTA \\
		\hline
		PoseTrack 17 val set & STEmbedding \cite{stembedding}$^*$ & 77.0 & 71.8 \\
		& {EOGN} \cite{offset_guided_networks} & 76.7 & 70.1\\
		& PGPT \cite{PGPT} & 77.2 & 68.4 \\
		& \textbf{Ours + Merge}  & \textbf{78.0} & \textbf{68.3} \\
		& \textbf{Ours} & \textbf{78.0} & \textbf{67.9} \\
		& POINet \cite{poinet} & - &  65.9 \\
		& HRNet \cite{hrnet}  & 77.3 & - \\
		& FlowTrack \cite{simple_baseline}  & 76.7 & 65.4 \\
		\hline
	 PoseTrack 17 test set & {EOGN} \cite{offset_guided_networks} & 74.8 & 61.1 \\
	 & PGPT \cite{PGPT} & 72.6 & 60.2 \\
	 &   \textbf{Ours + Merge} & \textbf{74.2} & \textbf{60.0} \\
	  &  POINet \cite{poinet}  & 72.5 & 58.4 \\
	  &  LightTrack \cite{ning2019lighttrack} &	66.8 &	58.0 \\
		&HRNet \cite{hrnet}& 75.0 & 58.0 \\ 
	&	FlowTrack  & 74.6 & 57.8 \\
	\end{tabular}}
    \end{minipage}\hfill%
    \begin{minipage}[t]{.49\linewidth}
     \centering
        \scalebox{0.70}{\begin{tabular}{l|llll}
        \hline
		 &  Approach & mAP & MOTA \\
		\hline
		PoseTrack 18 val set & 				\textbf{Ours + Merge} & \textbf{79.2} & \textbf{69.1} \\
		& \textbf{Ours}  & \textbf{79.2} & \textbf{68.8} \\
		&MIPAL \cite{mipal} & 74.6 &  65.7\\
		&LightTrack \cite{ning2019lighttrack} & 71.2 & 64.9 \\
		&Miracle$^+$ \cite{miracle} & 80.9 & 64.0 \\
		&OpenSVAI \cite{openSVAI}  & 69.7 & 62.4 \\
		&STAF \cite{staf}  & 70.4 & 60.9 \\
		& & \\
		\hline
PoseTrack 18 test set & MSRA$^+$ & 74.0 & 61.4 \\
		&ALG$^+$  & 74.9 & 60.8 \\
		&\textbf{Ours + Merge}  & \textbf{74.4} & \textbf{60.7} \\
	    &Miracle$^+$ \cite{miracle} & 70.9 &	57.4 \\
	    &MIPAL \cite{mipal} & 67.8 & 54.9 \\
	    &CV-Human  & 64.7 & 54.5 \\ 
	    && \\
	\end{tabular}}
    \end{minipage} 
\end{table}
We compare to the state-of-the-art for multi-frame pose estimation and multi-person pose tracking on the PoseTrack 2017 and 2018 datasets.
\paragraph{Multi-Frame Pose Estimation.}
For the task of multi-frame pose estimation, we compare to the state-of-the-art on the PoseTrack 2017 and 2018 validation sets, respectively. Although our correspondences are trained without using any video data, our approach outperforms the recently proposed PoseWrapper \cite{PoseWrapper}  approach on the PoseTrack 2018 validation set with mAP of ${82.0}$ and achieves very competitive mAP on the PoseTrack 2017 validation set with mAP of ${80.8}$ as shown in Table~\ref{table:multi_frame_pose_sota}.

\paragraph{Multi-Person Pose Tracking.}
We compare our tracking approach with the state-of-the-art for multi-person pose tracking on the PoseTrack 2017 and 2018 validation sets and leaderboards. In addition, we perform a post-processing step in which we merge broken tracks similar to the recovery of missed detections described in Section \ref{sec:pose_propagation}. This further improves the tracking performance. For further details we refer to the supplementary material.

We submitted our results to the PoseTrack 2017 and 2018 test servers, respectively. Our approach achieves top scoring MOTA of ${60.0}$ on the PoseTrack 2017 leaderboard without any bells and whistles as shown in Table \ref{table:state-of-the-art-tracking}. Our tracking performance is on-par with state-of-the-art approaches on the PoseTrack 2017 validation set. 

Similarly, we achieve top scoring MOTA of ${69.1}$ on the PoseTrack 2018 validation set as shown in Table \ref{table:state-of-the-art-tracking}. Our tracking results are very competitive to the winning entries on the PoseTrack 2018 leaderboard although the winning entries use additional training data.

\begin{figure*}[t]
\begin{center}
 \includegraphics[width=0.99\linewidth]{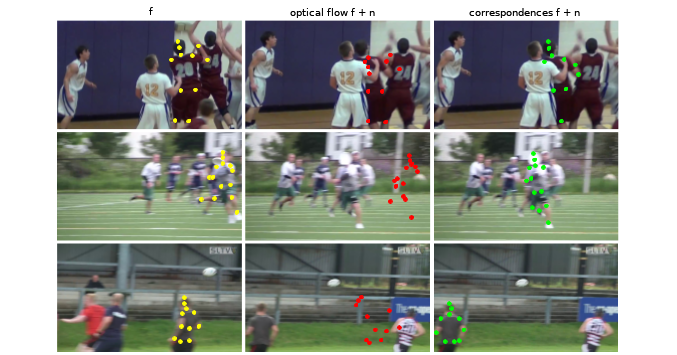}
\end{center}
\caption{Qualitative comparison between optical flow and correspondences for the task of pose warping under occlusion, motion blur, and large motion. (a) Query pose in frame $f$. (b) Warped pose using optical flow. (c) Warped pose using correspondences. In contrast to optical flow, the correspondences warp the poses correctly despite of occlusions, motion blur, or large motion.}
\label{fig:optical_flow_vs_corr}
\end{figure*}

\subsection{Qualitative Results}\label{sec:qualitative_results}
We qualitatively compare optical flow and correspondences for the task of pose warping under motion blur, occlusions, and large motion in Figure \ref{fig:optical_flow_vs_corr}. While the column on the left hand side shows the query pose in frame $f$. The columns in the middle and on the right hand side show the warped poses generated by optical flow or correspondences, respectively. In contrast to optical flow, our approach is robust to occlusion and fast human or camera motion. Our approach, however, has also some limitations. For instance, we observe that we obtain sometimes two tracks for the same person if the person detector provides two or more bounding boxes for a person like one bounding box for the upper body and one bounding box for the full body. Examples of failure cases are shown in the supplementary material.

\section{Conclusion}

In this work, we have proposed a self-supervised keypoint correspondence framework for the tasks of multi-frame pose estimation and multi-person pose tracking. 
The proposed keypoint correspondence framework solves two tasks: (1) recovering missed detections and (2) associating human poses across video frames for the task of multi-person pose tracking. The proposed approach based on keypoint correspondences outperforms the state-of-the-art for the tasks of multi-frame pose estimation and multi-person pose tracking on the PoseTrack 2017 and 2018 datasets.

\subsection*{Acknowledgment} The work has been funded by the Deutsche Forschungsgemeinschaft (DFG, German Research Foundation) GA 1927/8-1 and the ERC Starting Grant ARCA (677650).

%% file: include/supplementary_material.tex
\begin{appendix}

\section{Supplementary Material}
\subsection{Pose Estimation Framework}
\begin{figure}[h]
\begin{center}
 \includegraphics[width=1.\linewidth]{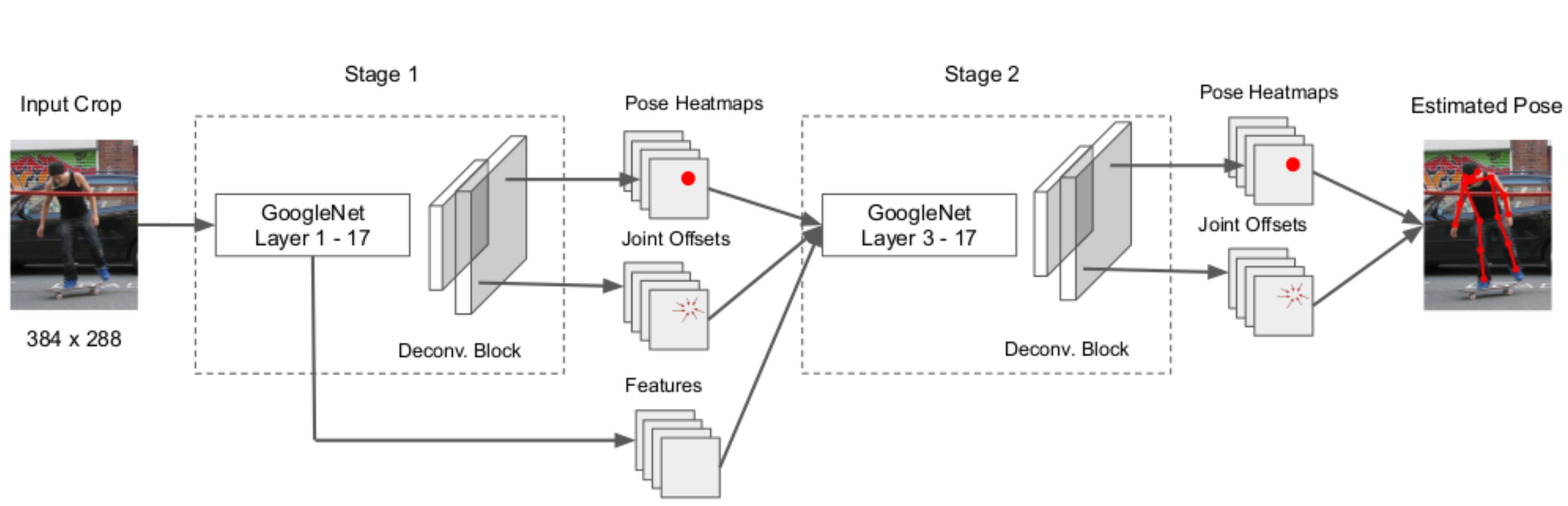}
\end{center}
 \caption{Our two-stage pose estimation framework.  Each stage uses GoogleNet \cite{googlenet} as the backbone. 
 The features extracted by the backbone in the first stage are fed into a deconvolution layer block to produce pose and joint offsets maps. The backbone features, pose heatmaps and joint offsets maps from the first stage are fed into the second stage to produce refined pose and joint offsets maps.}

\label{fig:pose_model}
\end{figure}
\noindent
Our two-stage pose estimation framework is shown in Figure \ref{fig:pose_model}. Each stage uses a GoogleNet \cite{googlenet} as the backbone. We use layer 1 to layer 17 for the backbone of the first stage while for the second stage we use layer 3 to layer 17 only.
The features extracted by the backbone in the first stage are fed into a deconvolution layer block to produce pose and joint offset maps. The backbone features, pose heatmaps and joint offset maps from the first stage are fed into the second stage to produce refined pose and joint offset maps.

Due to pooling used in the backbone, the resolution of the pose heatmaps is reduced by a factor of 4 in height and width dimensions. Consequently, the up-sampled predicted pose is slightly away from the actual pose. Towards this end, we append a joint offset head to predict the deltas, \ie, $ \Delta x$ and $\Delta y$ for each keypoint. The 
position of the $j$th keypoint $(\hat{x}_j, \hat{y}_j)$ at inference is computed as 
\begin{equation}
    (\hat{x}_j, \hat{y}_j) = (x_j  + 
    \Delta{x_j}, \;y_j + \Delta{y_j}).
\end{equation}
where $(x_j, y_j)$ is the up-sampled position from the pose heatmaps.
During training, we minimize the L1 loss between the predicted and ground-truth deltas for the joint offset maps and use the binary cross entropy loss for the pose heatmaps.

\subsection{Impact of $\tau_{corr}$}
We evaluate the impact of $\tau_{corr}$ on the pose estimation and tracking performance. As shown in Table \ref{table:hyperparam_eval}, the threshold has a low impact. We use $\tau_{corr} = 0.3$ for all our experiments.
\begin{table}[t]
\centering
	\caption{Impact of $\tau_{corr}$ on mAP and MOTA on the PoseTrack 2017 validation set.}\label{table:hyperparam_eval}
\begin{tabularx}{0.9\textwidth}{XXc}
\hline\noalign{\smallskip}
		 $\tau_{corr}$ & MOTA  & mAP\\
		\hline
		 0.1 & 67.9 & 77.9 \\
		 0.2 & 67.9 & 77.9 \\
		\textbf{0.3} & \textbf{67.9} & \textbf{78.0} \\
		 0.4 & 67.9 & 78.0 \\
		 0.5 & 67.8 & 78.0 \\
\end{tabularx}
\end{table}

\subsection{Effect of Refinement Module and Duplicate Removal}\label{supp:tracking}
\setlength{\tabcolsep}{4pt}
\begin{table}[t]
	\centering
		\caption{Comparison of mAP and MOTA for different design choices on the PoseTrack 2017 validation set.}\label{table:design_choices}
	\scalebox{1.}{\begin{tabular}{l|ll|l}
		\hline\noalign{\smallskip}
		Design Choices & MOTA  & mAP & IDSW\\
		
		\noalign{\smallskip}
		\hline
		\noalign{\smallskip}
	    Correspondence Tracking & 67.9 & 78.0 & 3632\\
	     Correspondence Tracking w/o refinement module & 66.9 & 77.7 & 4304  \\ 
		Correspondence Tracking w/o duplicate removal &  64.5 & 77.9 & 8288 \\
		\hline
	\end{tabular}}

\end{table}

We evaluate the effect of the refinement module and duplicate removal on the pose estimation and tracking performance.
  As shown in Table \ref{table:design_choices}, omitting any of the introduced design choices results in a significant drop in MOTA of at least $1\%$, and increases the number of identity switches (IDSW).
Our proposed correspondence refinement module improves the generated correspondence affinity maps which results in stronger tracking results. This is reflected by the MOTA and mAP scores that drop to $66.9$ and $77.7$, respectively, if we disable the refinement module.
If duplicates are not removed, the MOTA and the mAP scores drop to $64.5$ and $77.9$, respectively.

\subsection{Track Merging}\label{supp:merging}
\begin{figure}[h]
\begin{center}
 \includegraphics[width=0.9\linewidth]{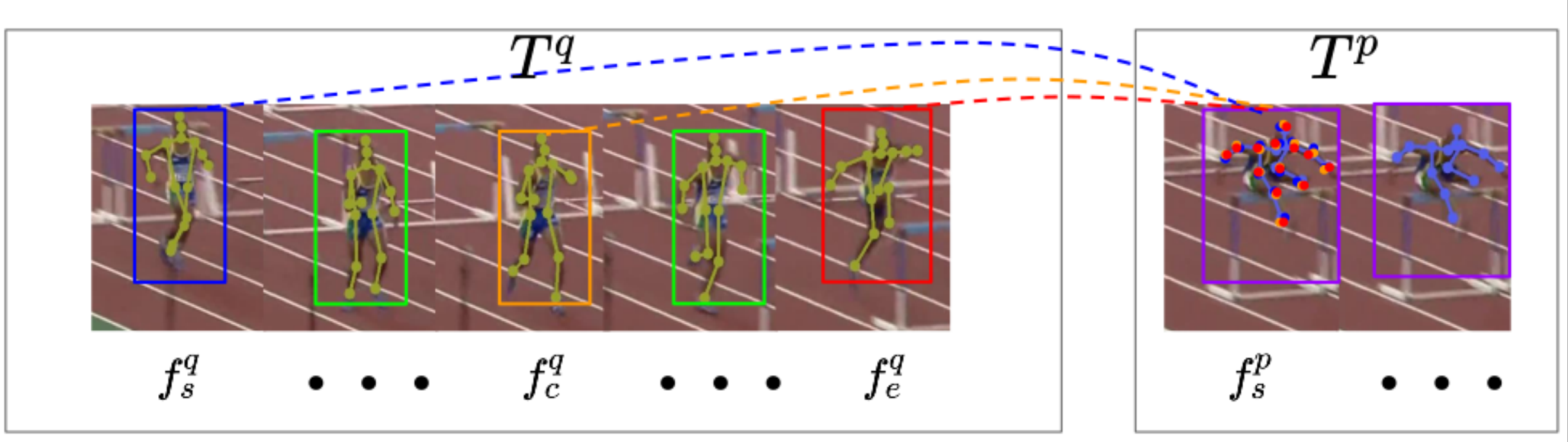}
\end{center}
 \caption{Tack merging: For the start frame $f_{s}^q$, the center frame $f_{c}^q$ and the last frame $f_{e}^q$ of track $T^q$, we estimate poses from keypoint correspondences in the start frame $f_{s}^p$ of $T^p$, as illustrated by the colored dashed lines. We use an OKS-based similarity metric to measure the average pose similarity between the poses from correspondences and the pose in the starting frame $f_s^p$ of track $T^p$.}
\label{fig:track_merging}
\end{figure}
We propose a post-processing step in which we merge tracks of the same pose instance at different time steps by utilizing keypoint correspondences from multiple frames. Given two tracks $T^q$ and $T^p$ as illustrated in Figure \ref{fig:track_merging}, we select three pose instances $\lbrace B^q_f \rbrace$ with $f \in \lbrace f_{s}^q, f_{c}^q, f_{e}^q\rbrace$ at the start, center and end frames of track $T^q$. For each of the pose instances $B^q_f$, we compute the pose $\bar{B}_f^q$ for the starting frame $f_s^p$ of track $T^p$ using correspondences, as described in Section 5 of the paper. We then employ OKS as similarity metric and calculate the average similarity between tracks $T^q$ and $T^p$ as 
\begin{equation}
S_{match}(T^q, T^p) = \frac{\sum_{f \in \lbrace f_s, f_c, f_e \rbrace} OKS(\bar{B}^q_f, B^p_{f_{s}^p})}{3}.
\end{equation}

\section{Failure Cases}
\begin{figure}[h]
\begin{center}
 \includegraphics[width=1.\linewidth]{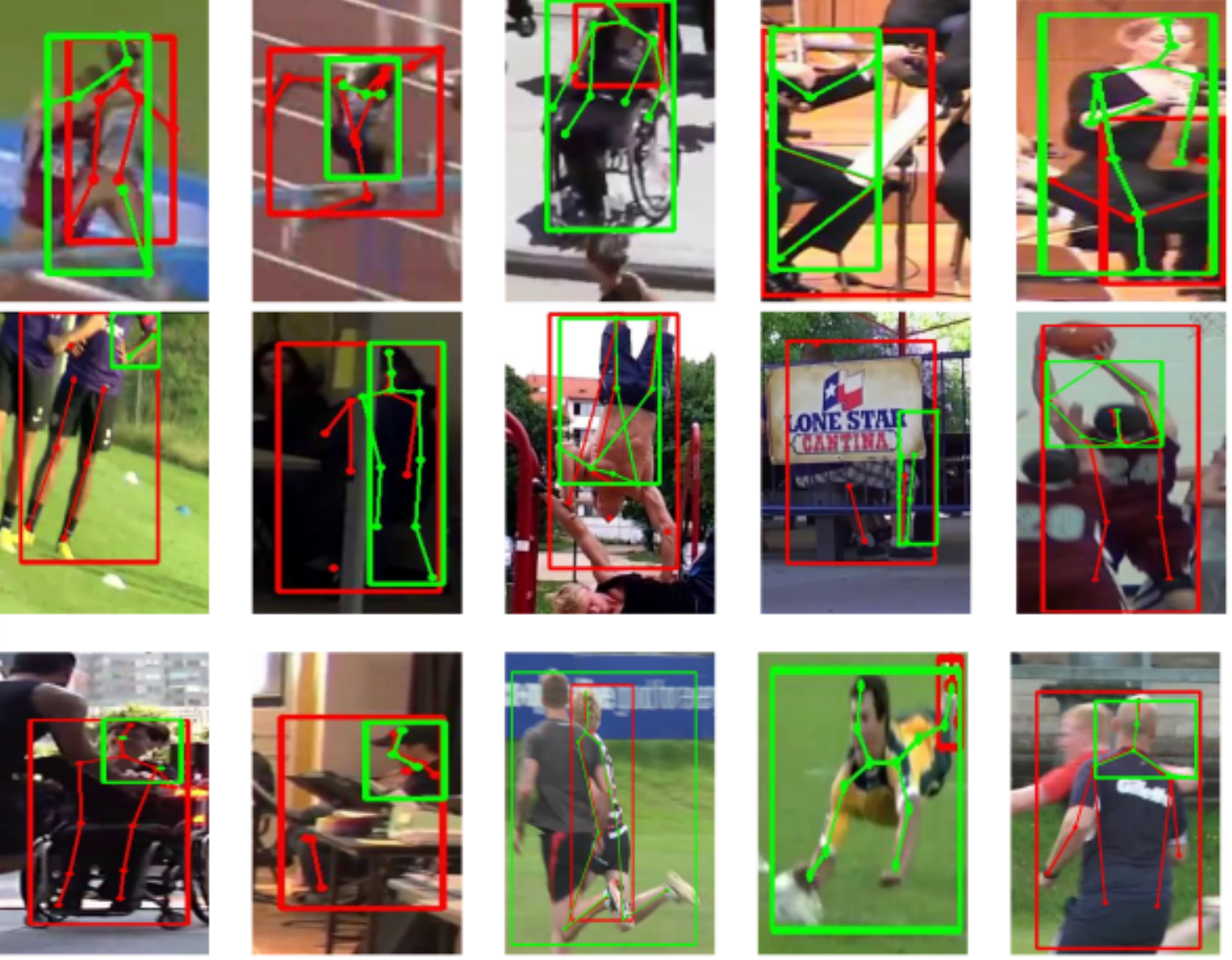}
\end{center}
 \caption{Failure cases. Duplicates by the person detector lead to multiple tracks of the same person and negatively impact the tracking performance.}
\label{fig:failure_cases}
\end{figure}
Existing person detectors sometimes output duplicate detections for the same person. Such duplicate detections are hard to remove using non-maximum suppression. In our experiments, they increase the number of false-positives (FP) and lead to identity-switches. This impacts the overall tracking performance, as the MOTA metric used in PoseTrack heavily penalizes FPs and IDSWs as shown in Table \ref{table:design_choices}. Figure \ref{fig:failure_cases} illustrates such failure cases.

\clearpage
\subsection{Qualitative Results}
\begin{figure}[h]
\begin{center}
 \includegraphics[width=0.97\linewidth]{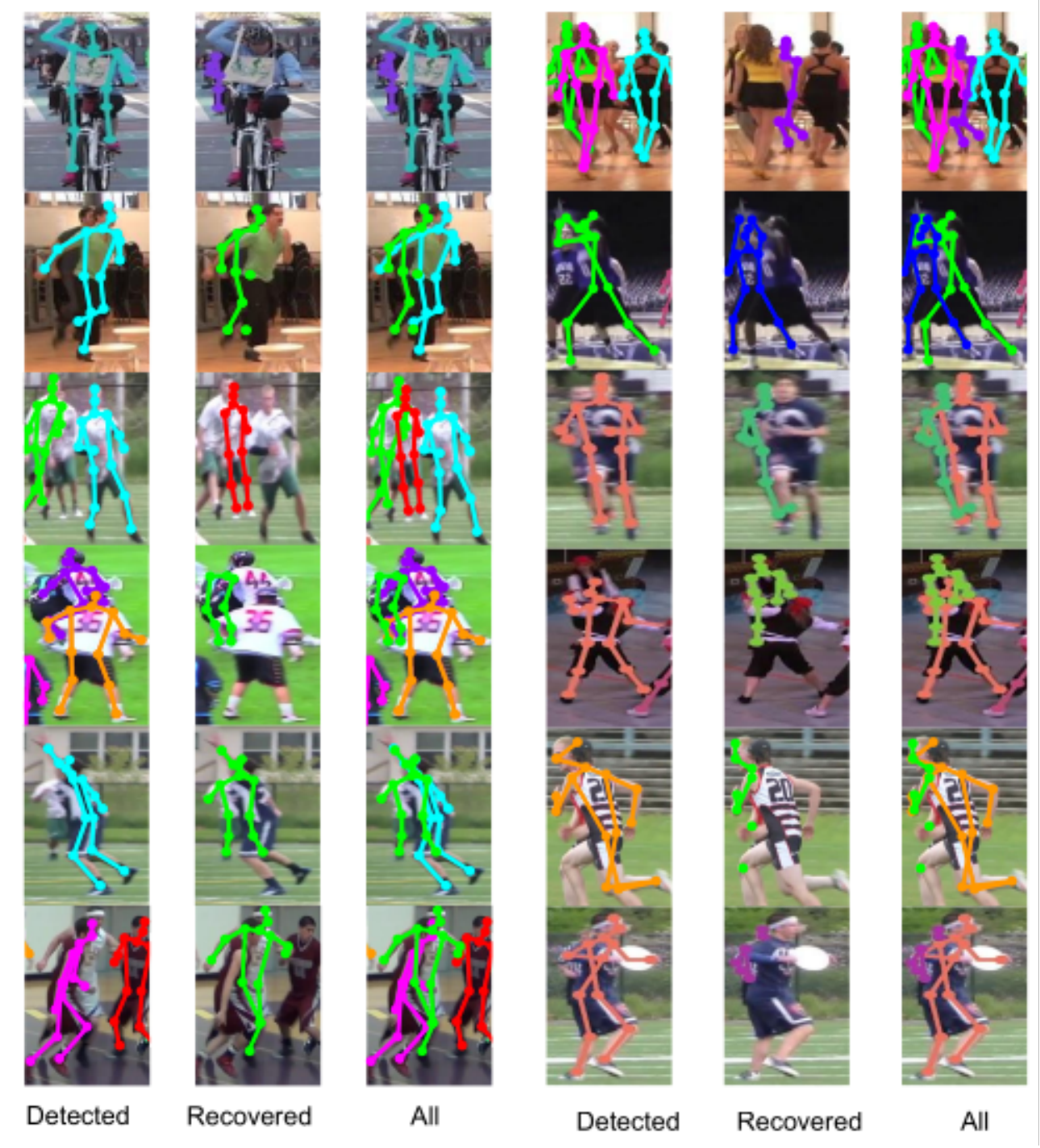}
\end{center}
 \caption{Qualitative results for recovering missed detections. Best seen using the zoom function of the PDF viewer.}
\label{fig:qualitative_pose_recovery}
\end{figure}

\end{appendix}